\documentclass[10pt,a4paper,conference]{IEEEtran}
\usepackage{color}
\usepackage{cite}

\usepackage{graphicx}
\usepackage{amsmath,amssymb}
\usepackage{algorithm}
\usepackage{algorithmicx}
\usepackage{algpseudocode}

\usepackage{array}

\usepackage[caption=false,font=normalsize]{subfig}
\usepackage{url}

\begin{document}
%
\title{Bottom-up Pose Estimation of Multiple Person \\with Bounding Box Constraint}



%
\author{\IEEEauthorblockN{Miaopeng Li,
Zimeng Zhou,
Jie Li,
Xinguo Liu}
\IEEEauthorblockA{State Key Lab of CAD \& CG, College of Computer Science and Technology, Zhejiang University}}


\maketitle
\begin{abstract}
In this work, we propose a new method for multi-person pose estimation which combines the traditional bottom-up and the top-down methods. Specifically, we perform the network feed-forwarding in a bottom-up manner, and then parse the poses with bounding box constraints in a top-down manner. In contrast to the previous top-down methods, our method is robust to bounding box shift and tightness. We extract features from an original image by a residual network and train the network to learn both the confidence maps of joints and the connection relationships between joints. During testing, the predicted confidence maps, the connection relationships and the bounding boxes are used to parse the poses of all persons. The experimental results showed that our method learns more accurate human poses especially in challenging situations and gains better time performance, compared with the bottom-up and the top-down methods.
\end{abstract}


%
\IEEEpeerreviewmaketitle

\section{Introduction}
Multi-person pose estimation aims to locate the skeletal keypoints for all persons in an image. Recognizing poses of multiple persons in the wild is very challenging due to the high flexibility of body limbs, self and outer occlusion, various clothing, challenging view condition, unusual pose, and etc. 

Significant progresses have been made on 2D human pose estimation, since deep convolutional neural networks (CNNs) are developed for this task. Existing approaches can be divided into two categories: the top-down approaches and the bottom-up approaches. The top-down approaches first obtain human candidates by a human detector, and then perform single-person pose estimation. While the bottom-up approaches directly predict the keypoints all at once, and then assemble them into the full poses for all persons. 

\begin{figure}
	\centering
	\subfloat{\rotatebox{90}{\footnotesize ~~Top-down}}\
	\subfloat{\includegraphics[height=16.2mm]{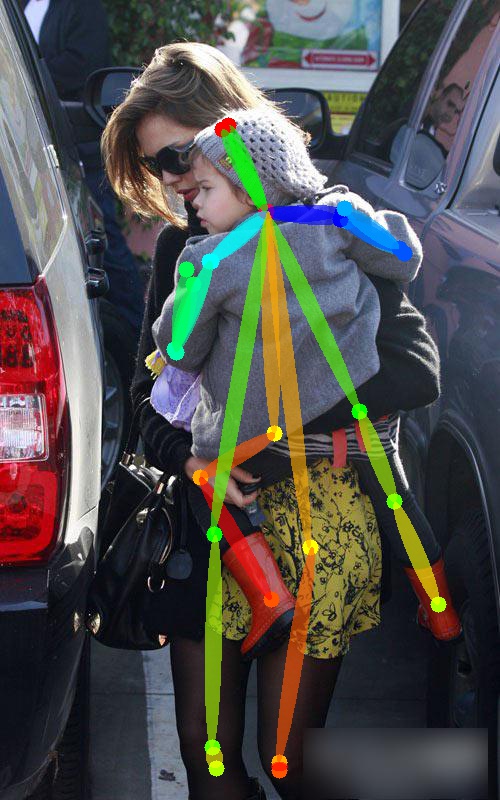}}\
	\subfloat{\includegraphics[height=16.2mm]{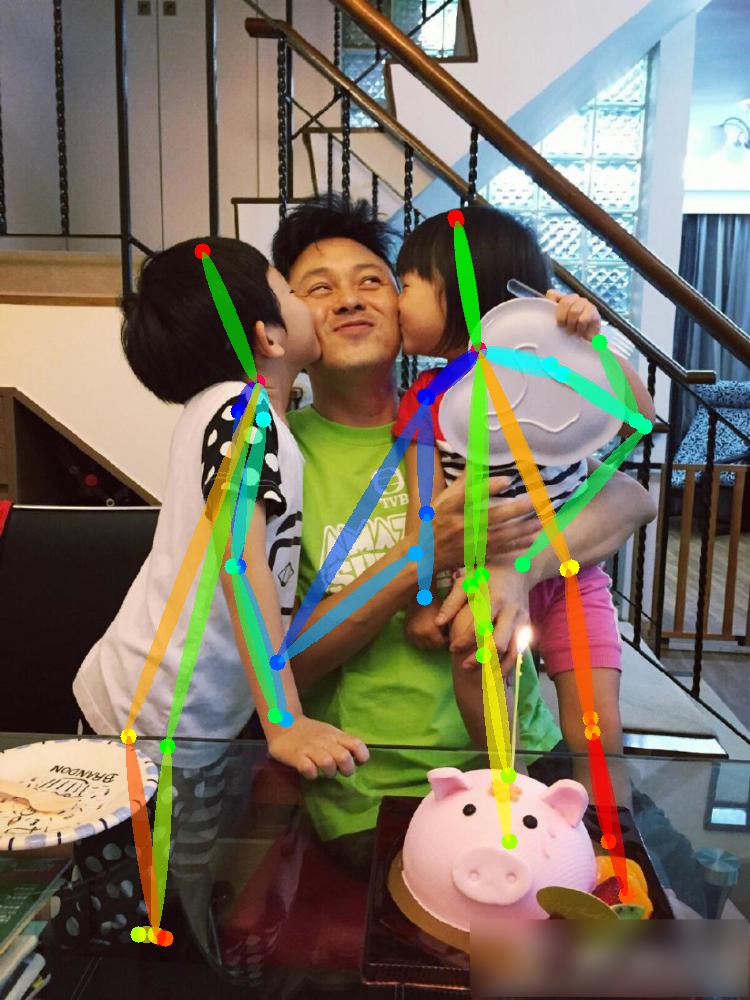}}\
	\subfloat{\includegraphics[height=16.2mm]{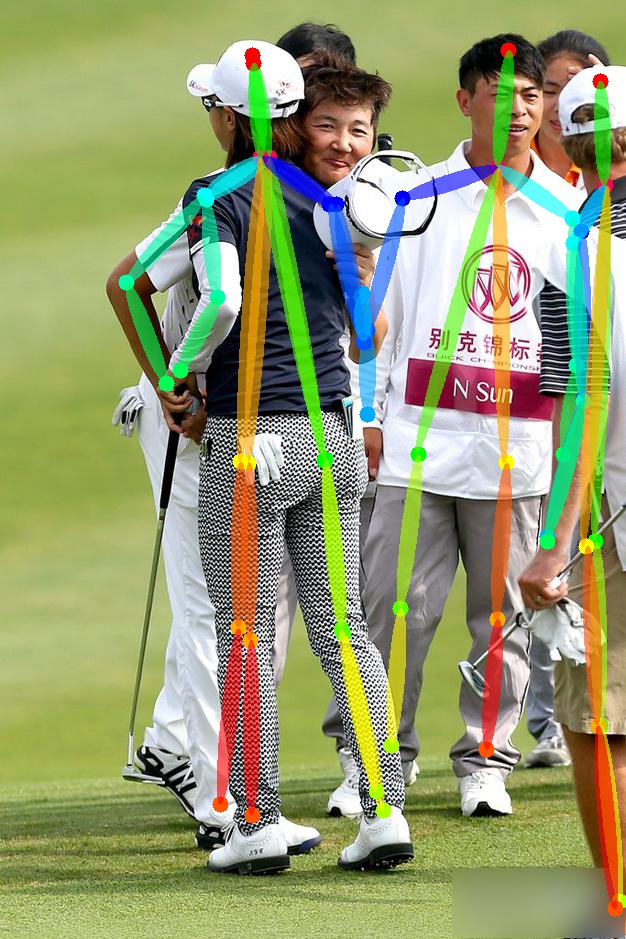}}\
	\subfloat{\includegraphics[height=16.2mm]{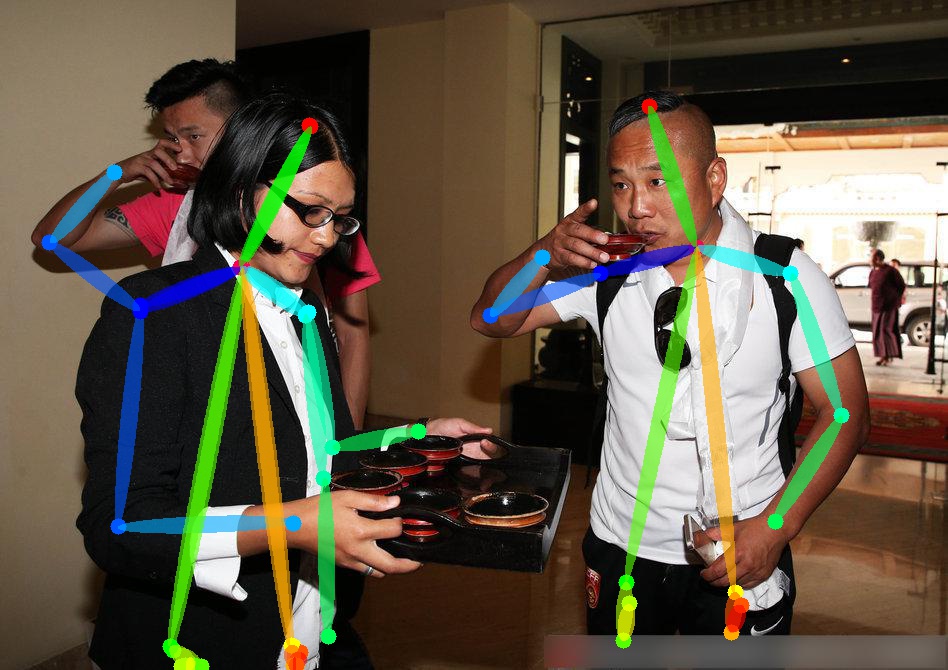}}\
	\subfloat{\includegraphics[height=16.2mm]{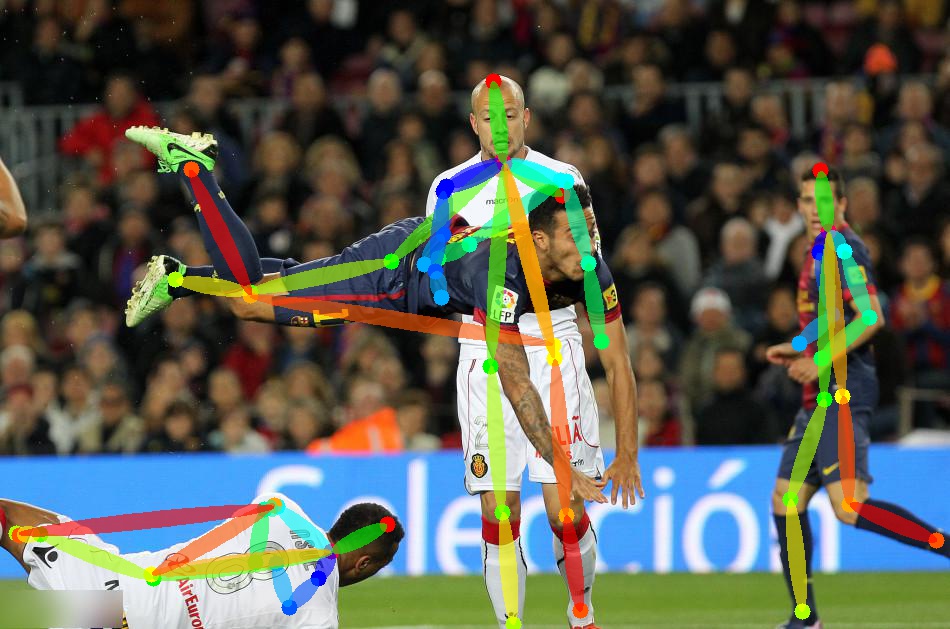}}\\
	\vspace{1mm}
	\subfloat{\rotatebox{90}{\footnotesize ~~Bottom-up}}\
	\subfloat{\includegraphics[height=16.2mm]{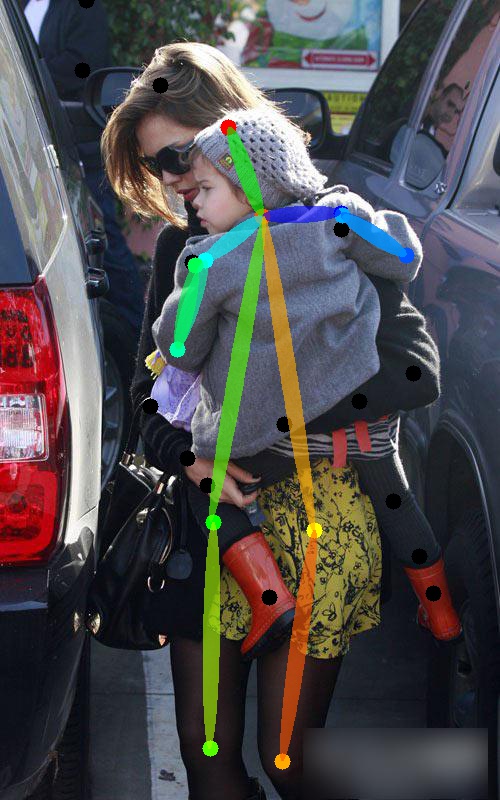}}\
	\subfloat{\includegraphics[height=16.2mm]{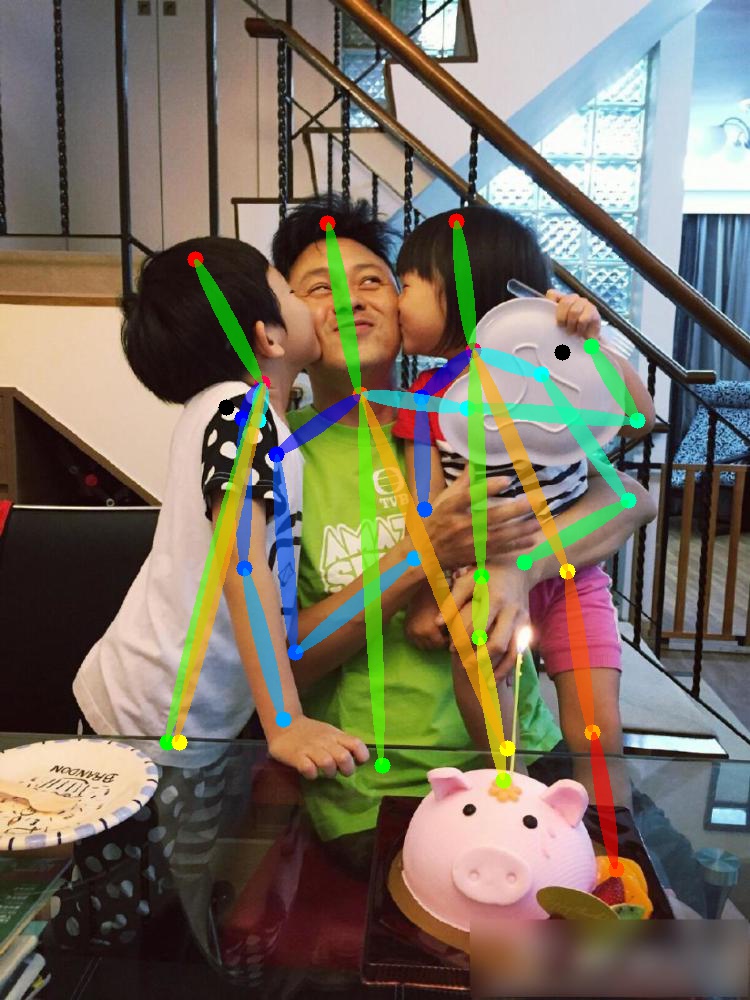}}\
	\subfloat{\includegraphics[height=16.2mm]{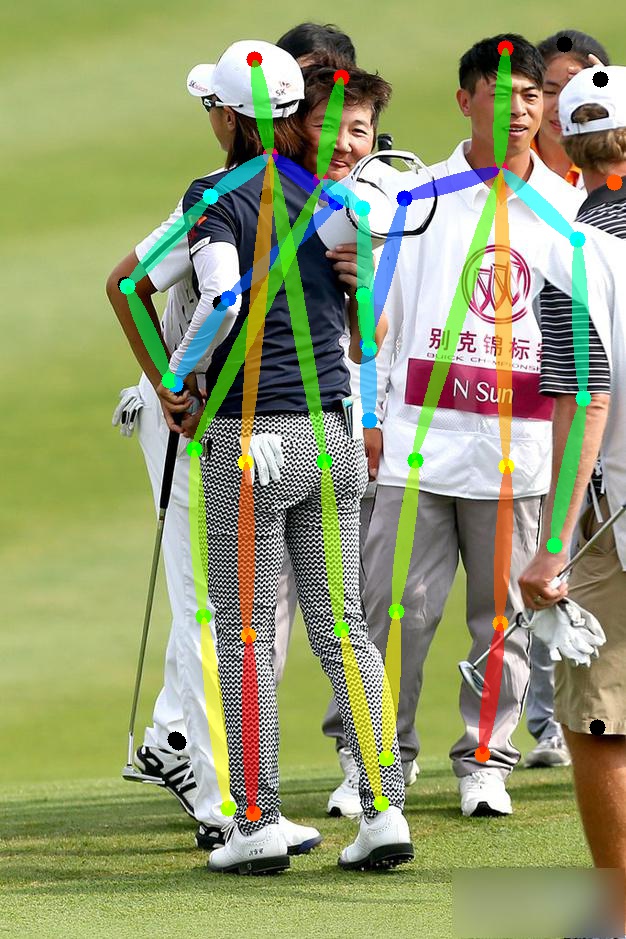}}\
	\subfloat{\includegraphics[height=16.2mm]{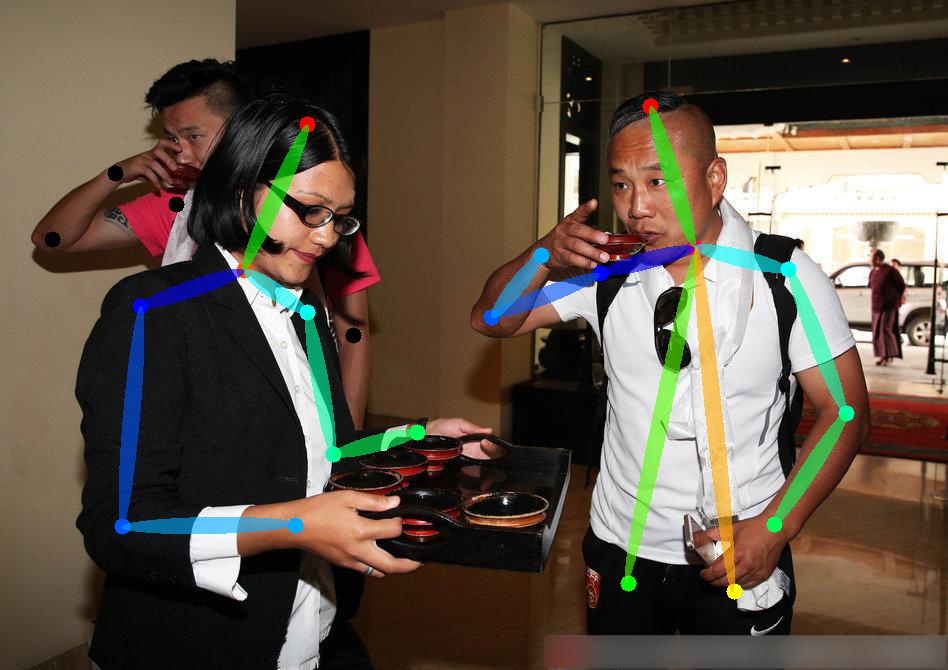}}\
	\subfloat{\includegraphics[height=16.2mm]{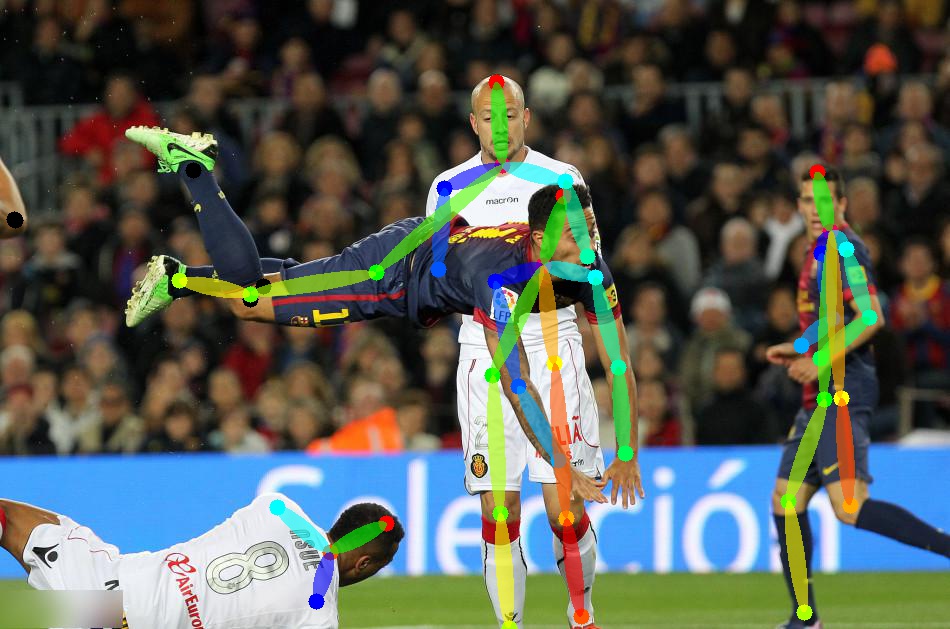}}\\
	\vspace{1mm}
	\setcounter{subfigure}{-1}
	\subfloat{\rotatebox{90}{\footnotesize ~~~~~Ours\textcolor{white}{g}}}\
	\subfloat[]{\includegraphics[height=16.2mm]{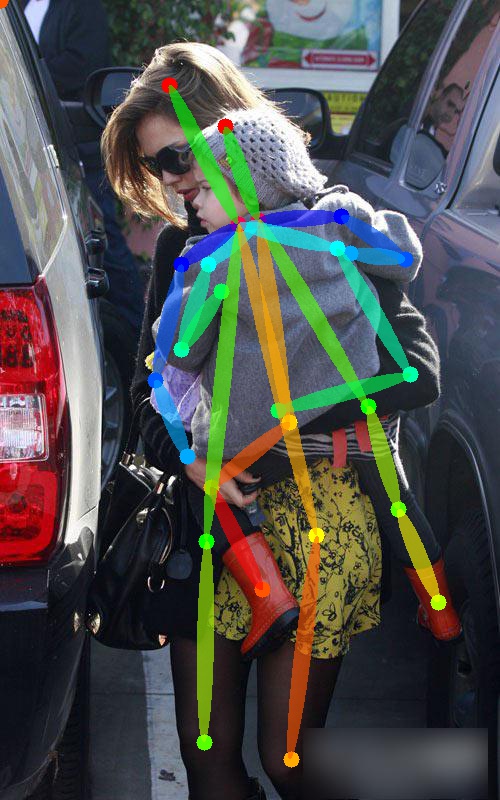}}\
	\subfloat[]{\includegraphics[height=16.2mm]{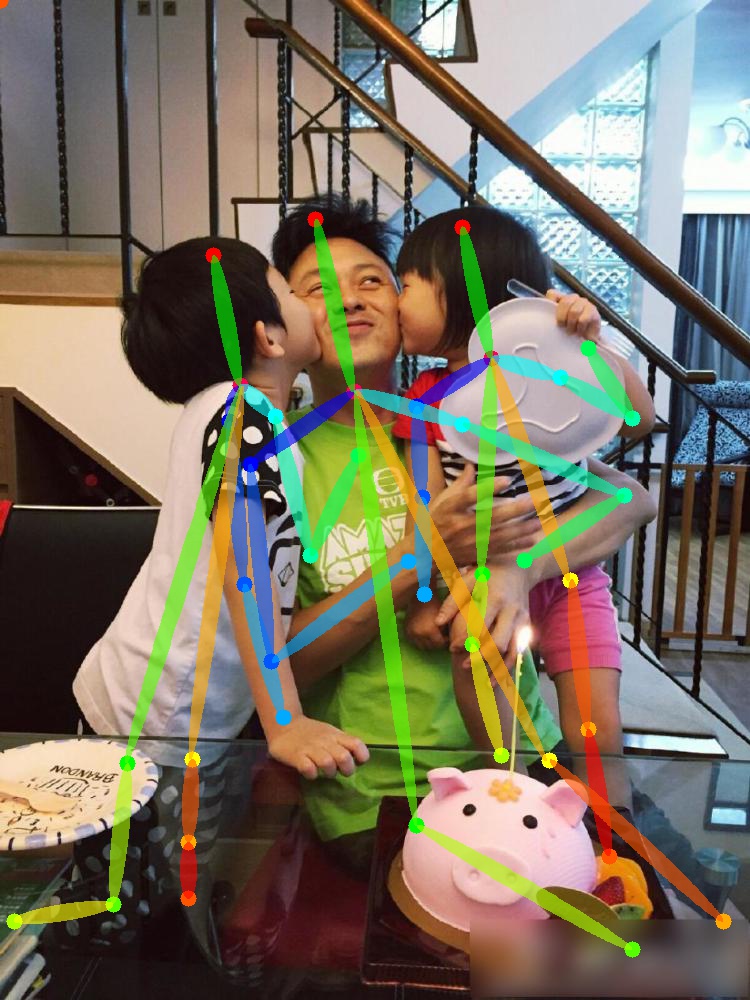}}\
	\subfloat[]{\includegraphics[height=16.2mm]{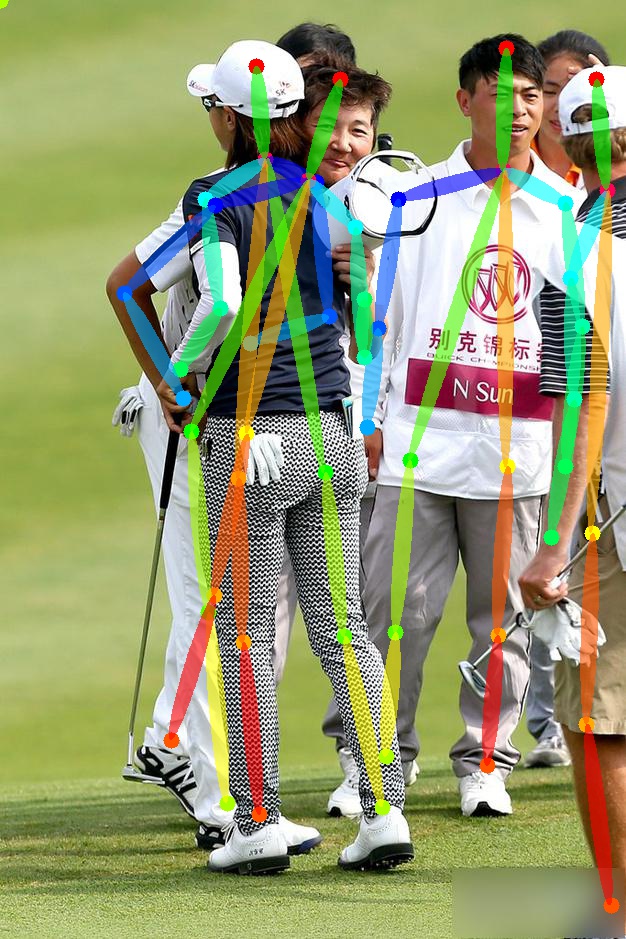}}\
	\subfloat[]{\includegraphics[height=16.2mm]{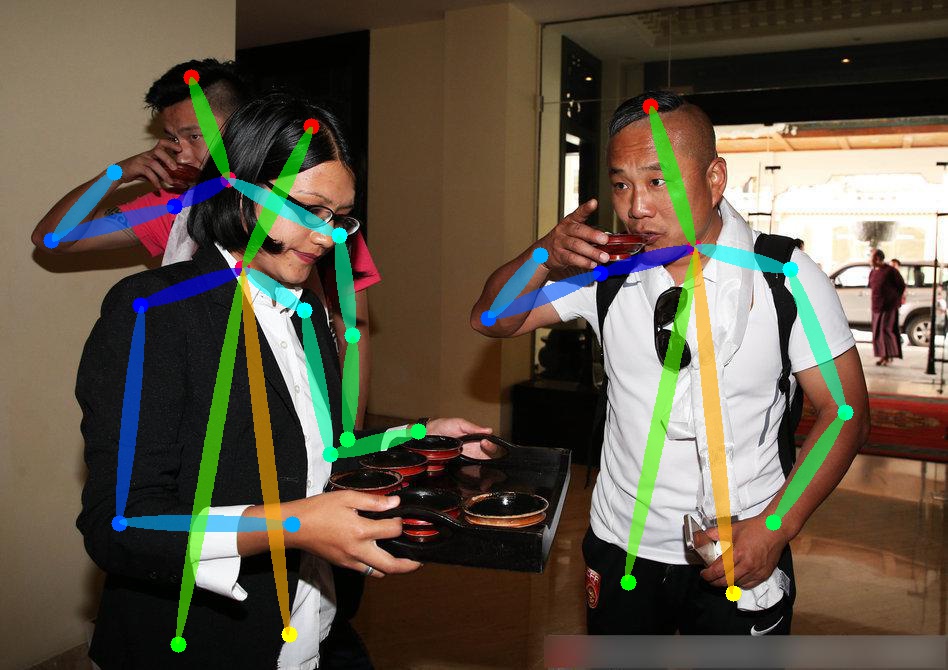}}\
	\subfloat[]{\includegraphics[height=16.2mm]{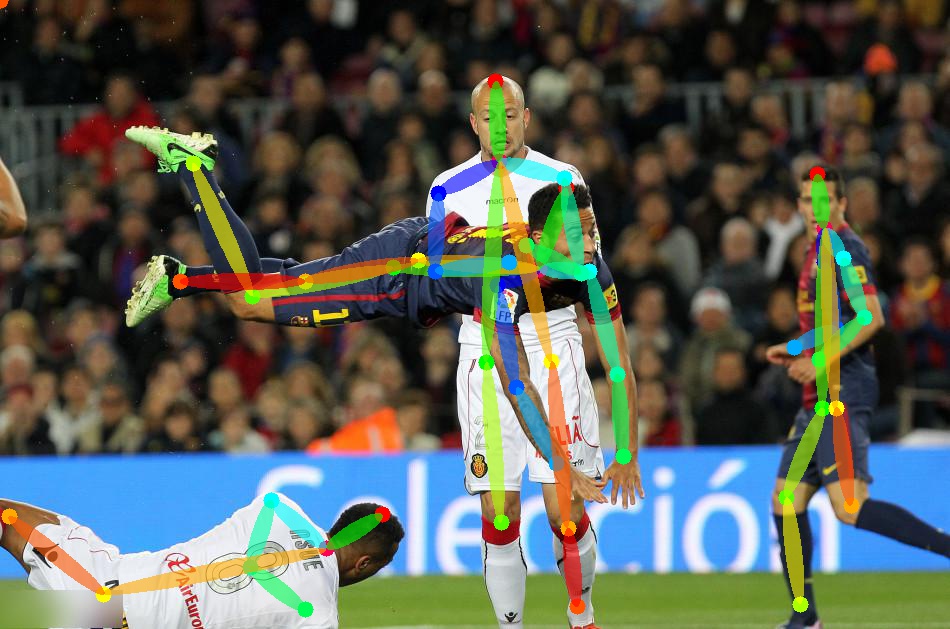}}\\
	
	\caption{Pose estimation results comparision. The first row: results from the top-down method using CPM \cite{wei2016convolutional} as the single-person pose estimator. The second row: results from the bottom-up method proposed by Cao et al. \cite{cao2016realtime}. The third row: results from our method.}
	\label{fig:comparision}
	\vspace{-4mm}
\end{figure}

Both categories have some shortcomings. The top-down approaches are very sensitive to bounding box shift and tightness (Fig.\,\ref{fig:comparision}\,(a-b)), because their model is usually trained with person-centered image patches. And they cannot handle overlapped persons well (Fig.\,\ref{fig:comparision}\,(c-d)), because it is too confusing for the network to determine which person should be annotated. Thirdly, they are time-consuming because the pose of each person is estimated independently and the inference time is proportional to the number of detected persons. 

On the contrary, the bottom-up methods are efficient in computing the joint locations\cite{pishchulin2016deepcut,insafutdinov2016deepercut}, but it takes long time to connect the corresponding joints for the persons. Cao et al. \cite{cao2016realtime} propose a greedy method for fast connecting the joints. However, it may produce disconnected parts of pose due to truncation or heavy occlusion (Fig.\,\ref{fig:comparision}\,(a)\,(c-d)). And, wrongly connected joints may propagate across the poses of different persons (Fig.\,\ref{fig:comparision}\,(b-c)\,(e)). 

In this paper, we aim to address the above problems, and improve the accuracy and the runtime speed. 
We propose a new approach combining the bottom-up pipeline with human detection. At the training stage, the network is trained to learn the keypoints of all persons indiscriminately. At the testing stage, we perform the network feed-forwarding for the whole image only once, then parse and refine the pose of each person within their bounding boxes. The bounding box constraint narrows down the scope of possible connections and avoids mistake propagation across the poses of different persons.

The contributions in this paper are: (i) we propose a simple yet effective method for multi-person pose estimation; (ii) we design an efficient multi-stage residual network to learn the confidence maps of the joints and the connection relationships between the joints; (iii) we test our method on a number of benchmark datasets, including AI Challenger and MSCOCO with a detailed analysis on the functions of different components in our design, and the experimental results showed the improved accuracy (Table\,\ref{table:result_AIC_val},\,\ref{table:result_COCO_val}) and runtime speed (Table\,\ref{table:dilation_models_AIC_val}). Please see our video demo at \url{https://youtu.be/J58hdvz7_SM} for vivid and intuitive visualization of results.

\section{Related Work}
Human pose estimation is a very challenging problem. The traditional approaches to this problem mainly adopt the techniques of pictorial structures \cite{fischler1973representation,andriluka2009pictorial} or graphical models \cite{chen2014articulated}. 
Recent works \cite{toshev2014deeppose,newell2016stacked, 
bulat2016human,wei2016convolutional} based on CNNs have improved the performance significantly. 
In this paper, we mainly focus on the methods based on CNNs. 

Human pose estimation can be categorized into single-person pose estimation, which predicts the single pose in a person-centered image,  and multi-person pose estimation, which predicts the poses of all persons in an image. In contrast to the single-person pose estimation, multi-person pose estimation is more challenging due to inter-occlusion, various scales of individual persons and unpredictable interactions between different persons.  Existing approaches to multi-person pose estimation can be divided into two groups: the top-down approaches and the bottom-up approaches.

{\bfseries Top-down approaches.}
The top-down approaches\cite{papandreou2017towards,fang2016rmpe} obtain human candidates by a human detector, and then perform single-person pose estimation to predict human joints. 
Papandreou et al.\cite{papandreou2017towards} use the Faster-RCNN\cite{ren2015faster} as their human detector, and then predict both heatmaps and offset fields by a CNN to obtain the locations of joints. 
Hao et al.\cite{fang2016rmpe} propose a regional multi-person pose estimation (RMPE) framework, which facilitates pose estimation in the inaccurate human bounding boxes by introducing more components into their pipeline to refine the detection and pose estimation results. 
The accuracy of top-down approaches highly depends on the quality of the human detection results, because the single-person pose estimator is usually sensitive to the detected bounding boxes. 
Furthermore, the runtime of the top-down approaches is proportional to the number of detected people, it is time-consuming.

{\bfseries Bottom-up approaches.}
Bottom-up approaches \cite{cao2016realtime,iqbal2016multi,newell2017associative,pishchulin2016deepcut,insafutdinov2016deepercut} detect the human joints of all persons, and then assemble these joints into the poses for each person based on various joints association techniques. 
For example, Deepcut\cite{pishchulin2016deepcut}, proposed by Pishchulin et al., jointly labels candidate joints, and then assigns them to individual persons by integer linear programming (ILP). However, solving the ILP problem over a fully connected graph is a NP-hard problem and very time-consuming. 
Insafudinov et al. propose Deepercut\cite{insafutdinov2016deepercut} built on \cite{pishchulin2016deepcut} with stronger joint detectors and image-dependent pairwise scores, and vastly improve the runtime performance. But the method still takes several minutes per image. 
Cao et al.\cite{cao2016realtime} predict the keypoint heatmaps and part affinity fields (PAFs) of all persons, then group the candidate joints into each person by a greedy algorithm to gain a real-time performance. However, as shown in Fig.\,\ref{fig:comparision}, 
it  may produce disconnected parts of pose and propagate mistakes across different persons.

Motivated by the works in above, we propose a bottom-up approach to multi-person pose estimation with bounding box constraint, aiming to design  a more efficient network for joints detection and a more robust algorithm for joints association. 
\begin{figure}
	\centering
	\includegraphics[width=0.495\textwidth]{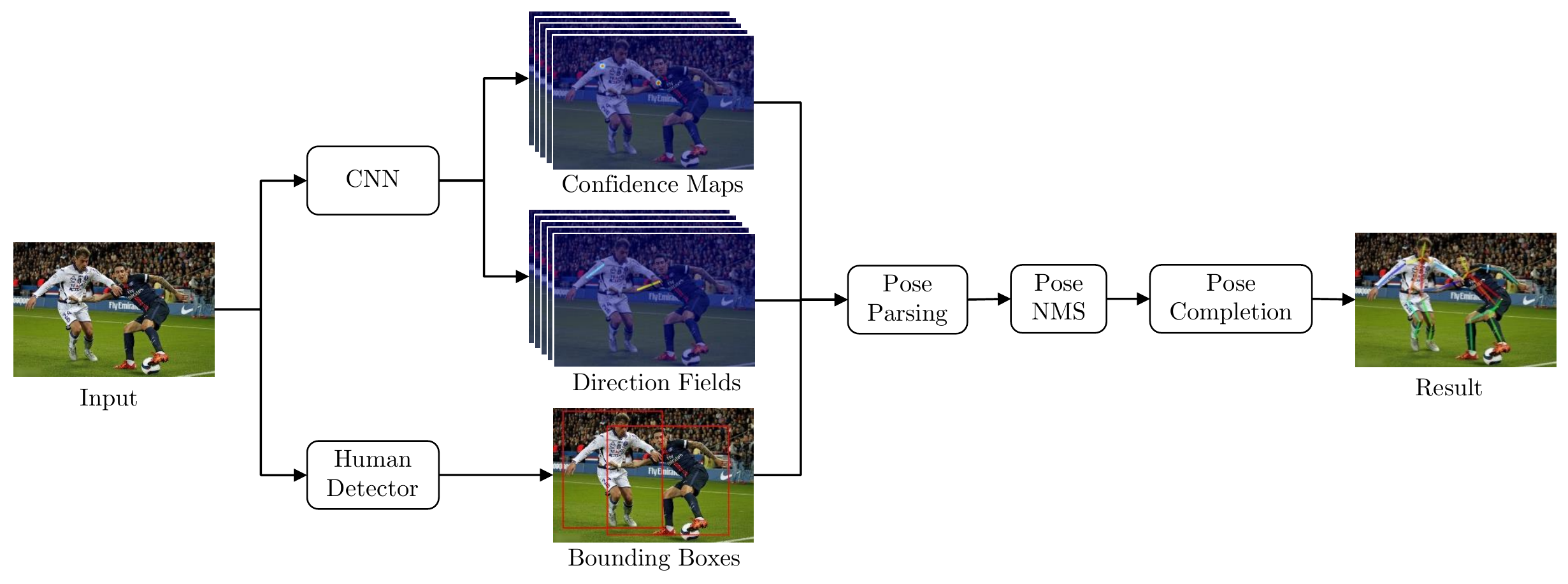}
	\caption{Overview. Given an image, firstly, estimate the confidence maps and the direction fileds by the trained CNN. Secondly, parse the pose for each person within corresponding bounding box. Thirdly, remove the redundant poses by pose NMS. Finally, complete the pose by associating the disconnected points.}
	\label{fig:pipeline}
\end{figure}
\section{Method}
\label{(sec:method)}
An overview of our approach is illustrated in Fig.\,\ref{fig:pipeline}. 
The input of our system is a color image $\mathbf{I}$, and the output is the 2D poses for all persons in the image. The output poses are represented by the 2D locations of a set of skeletal keypoints (head, neck, and etc., as shown in Fig.\,\ref{fig:skeleton}). In this section, we describe the four components in our approach in detail.

\subsection{CNN Regression}
The first component, CNN regression, generates the confidence maps $\mathbf{S}=(\mathbf{S}_1,\mathbf{S}_2,\ldots,\mathbf{S}_J)$ and the direction fields  $\mathbf{L}=(\mathbf{L}_1,\mathbf{L}_2,\ldots,\mathbf{L}_C)$. $\mathbf{S}_j \in \mathbb{R}^{w \times h}$, $j \in \{1,\ldots,J\}$ , one per joint, is a 2D representation of the confidence score that the $j\textrm{-th}$ joint occurs at each pixel location in image $\mathbf{I}$. $\mathbf{L}_c \in \mathbb{R}^{w \times h \times 2}$, $c \in \{1,\ldots,C\}$, one per limb, encodes a 2D vector field representing the connection relationship between two joints.
\begin{figure*}
	\centering
	\includegraphics[width=\textwidth]{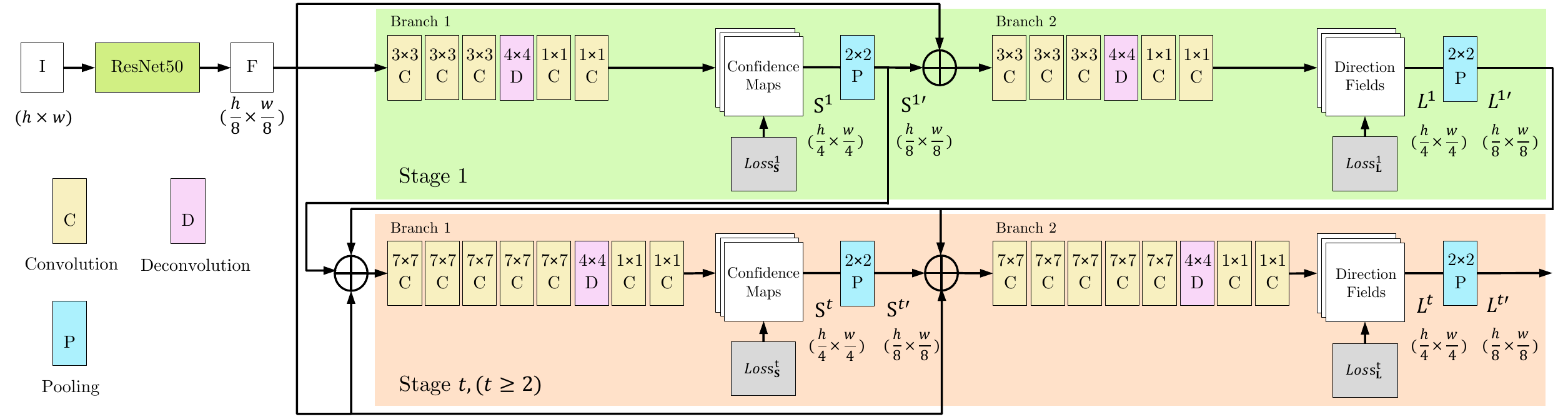}
	\caption{Structure of the two-branch multi-stage network. At each stage, the first branch predicts confidence maps $\mathbf{S}^t$, and the second branch predicts direction fields $\mathbf{L}^t$. The predictions from the two branches after downsampling, along with the image features $\mathbf{F}$, are concatenated for next stage.}
	\label{fig:network_architecture}
\end{figure*}
\subsubsection{Network Structure}
The network structure is shown in Fig.\,\ref{fig:network_architecture}, following the general network structure of Cao et al.\cite{cao2016realtime}, but with several modifications. Firstly, we use the ResNet50 \cite{he2016deep} with the hole algorithm \cite{chen2016deeplab} to replace the VGG19\cite{simonyan2014very} as the feature extractor. Secondly, we introduce deconvolution layers with $stride=2$ to upsample the outputs by 2 times, without incurring significant extra computation cost. Thirdly, at each stage, the confidence maps and the direction fields are predicted sequentially rather than concurrently, so as to further exploit spatial interdependencies. In the experiments, we find that a 3-stage network can obtain good performance in terms of both accuracy and speed (Table \ref{table:dilation_models_AIC_val}).

\subsubsection{Confidence maps for joints detection}
\label{sec:4.1.2}
The confidence map of a joint measures the probabilities of observing the joint at different locations. The ideal confidence map can be created by putting a Gaussian peak at the location of the joint. Let $\mathbf{S}^*_{j,k}$ represent the confidence map of joint $j$ for person $k$, and $\mathbf{x}_{j,k}$  the position coordinate of joint $j$. The confidence value at location $\mathbf{p}$ is defined as
\vspace{-1mm}
\begin{equation}
\mathbf{S}^*_{j,k}(\mathbf{p}) = \exp ( -{\|\mathbf{p}-\mathbf{x}_{j,k}\|} ^2 / \sigma^2 ),
\vspace{-1mm}
\end{equation}
where $\sigma$ controls the spread of the Gaussian peak. 

The confidence map  of joint $j$ for an input image with multiple persons is defined as 
\vspace{-1mm}
\begin{equation}
\mathbf{S}^*_{j}(\mathbf{p}) = \max_k \mathbf{S}^*_{j,k}(\mathbf{p}).
\vspace{-1mm}
\end{equation}

During testing, the confidence maps $\mathbf{S}$ are used to obtain the candidate joints by performing the non-maximum suppression.

\begin{figure}[tbp]
    \centering
    \includegraphics[width=0.42\textwidth]{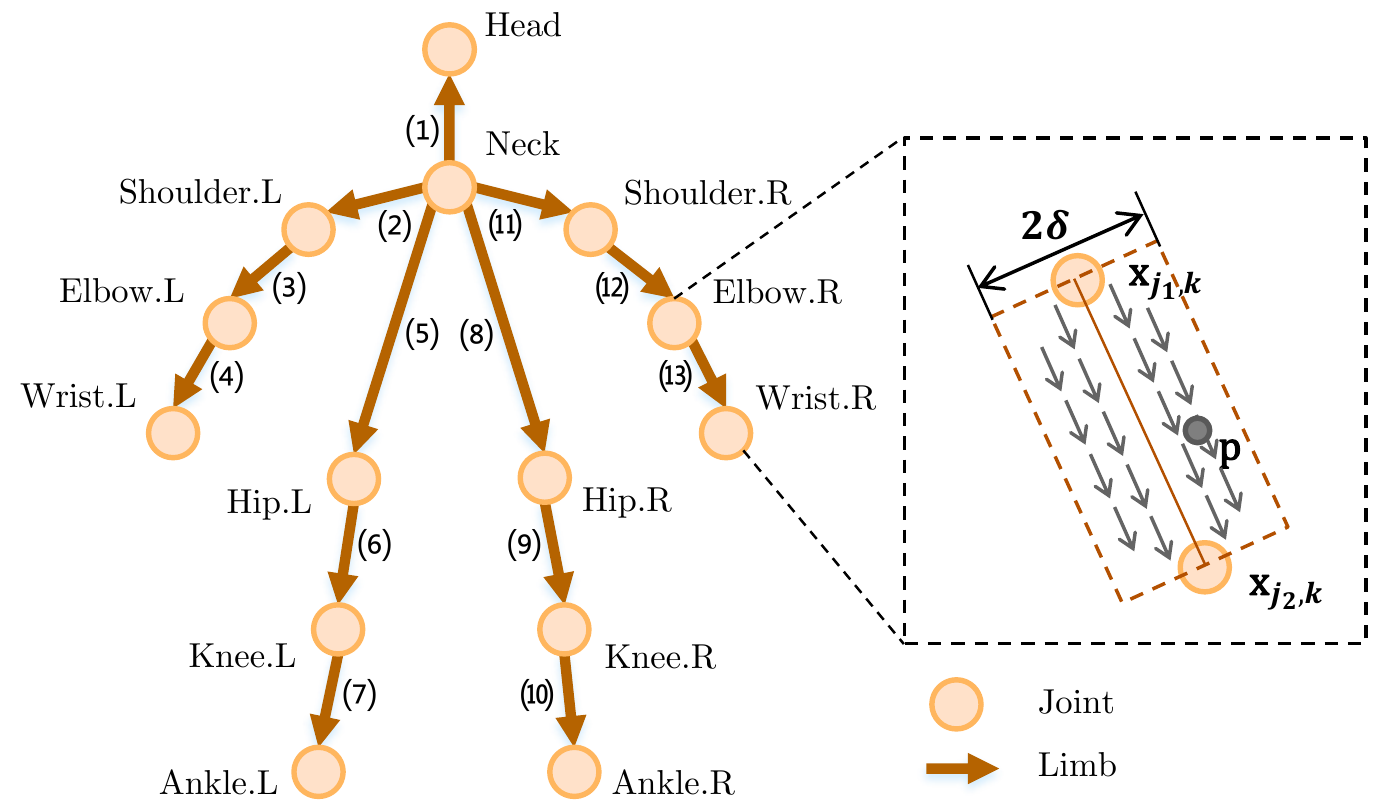}
    \caption{The human skeleton tree composed of 14 joints and 13 limbs. Neck is taken as the root. All limbs are ordered by Depth-First-Search (DFS).}
    \label{fig:skeleton}
    \vspace{-2mm}
\end{figure}
\subsubsection{Direction fields for joints association}
As shown in Fig.\,\ref{fig:skeleton}, there are 13 primary connections associating 14 joints, forming a tree structure. The direction fields encode the connection relationships between the joints. Each connection $c$ of person $k$ corresponds to a field $\mathbf{L}_{c,k}^*$, which is a 2D vector field encoding the position and the orientation information. As shown in Fig.~\ref{fig:skeleton}, the direction field is defined as vector $\mathbf{v}$ in a $2\delta$-wide rectangle. Then the value of direction field $\mathbf{L}_{c,k}^*$ at the point $\mathbf{p}$ is:
\vspace{-1mm}
\begin{equation}
\mathbf{L}_{c,k}^*(\mathbf{p}) = \left\{\begin{array}{ll}
\mathbf{v} & \quad \textrm{if $\mathbf{p}$ lies in the $2\delta$-wide rectangle,} \\
\mathbf{0} & \quad \textrm{otherwise.}
\end{array}\right.
\vspace{-1mm}
\end{equation}
where $\mathbf{v} = (\mathbf{x}_{j_2,k} - \mathbf{x}_{j_1,k})/\|\mathbf{x}_{j_2,k} - \mathbf{x}_{j_1,k}\|$ is the unit vector in the direction of connection $c$. 

The direction field of connection $c$ for an input image with multiple persons is defined as
\vspace{-1mm}
\begin{equation}
\mathbf{L}_{c}^*(\mathbf{p}) = \frac{1}{n_c(\mathbf{p})} \sum_k \mathbf{L}_{c,k}^*(\mathbf{p}),
\vspace{-1mm}
\end{equation}
where $n_c(\mathbf{p})$ is the number of non-zero vectors at point $\mathbf{p}$.

During testing, the direction fields $\mathbf{L}$ are used to associate the candidate joints. For every two possible connected joints $j_1$ and $j_2$
, we measure the confidence that $j_1$ and $j_2$ yield a limb connection. Specifically, let $\mathbf{x}_{j_1}$ and $\mathbf{x}_{j_2}$ be the location coordinates of  $j_1$ and $j_2$, and we define the reference direction as $\mathbf{d}=\mathbf{x}_{j_2}-\mathbf{x}_{j_1}$. We uniformly sample some points along the line segment connecting $j_1$ and $j_2$, yielding a sample point set $Q$. The confidence score for this connection is computed by the following cosine similarity between the direction field $\mathbf{L}_{c}$  and the reference direction $\mathbf{d}$,
\begin{equation}
\label{equation:score_connection}
s(j_1,j_2) = \frac{1}{|Q|} \sum_{\mathbf{q} \in Q} \mathbf{L}_c(\mathbf{q}) \cdot \frac{\mathbf{d}}{{\|\mathbf{d}\|}},
\end{equation}
where $|Q|$ is the number of sampled points, $|Q| \propto {\|\mathbf{d}\|}$.

\subsubsection{Loss term}
We apply supervision at each stage to address the vanishing gradients problem \cite{glorot2010understanding}. Meanwhile, we use a weight loss function to avoid penalizing the true positive predictions during training. What's more, 
as done in \cite{pinheiro2014recurrent}, we weight the gradient responses between foreground and background heatmap pixels, so that they equally contribute to the parameters update. Consequently, the loss at stage $t$ is
\vspace{-1mm}
\begin{equation*}
Loss^t_{\mathbf{S}} = \sum_{j} {\sum_\mathbf{p}{\mathbf{W}(\mathbf{p})\cdot\|\mathbf{S}_j^t(\mathbf{p})-\mathbf{S}^*_j(\mathbf{p})\|^2}} 
\end{equation*}
\vspace{-2mm}
\begin{equation}
+\lambda\sum_\mathbf{p}{\|\mathbf{S}_{B}^t(\mathbf{p})-\mathbf{S}^*_{B}(\mathbf{p})\|^2},
\end{equation}
\vspace{-2mm}
\begin{equation}
Loss^t_{\mathbf{L}} = \sum_{c} {\sum_\mathbf{p}{\mathbf{W}(\mathbf{p})\cdot\|\mathbf{L}_c^t(\mathbf{p})-\mathbf{L}^*_c(\mathbf{p})\|^2}},
\vspace{-1mm}
\end{equation}
where $\mathbf{S}^*_j$ is the ground-truth confidence map for joint $j$ and $\mathbf{S}^*_{B}$ for the background. $\mathbf{L}^*_c$ is the ground-truth direction field for connection $c$. $\mathbf{W}$ is a binary mask with $\mathbf{W}(\mathbf{p})=0$ when the annotation is missing at location $\mathbf{p}$. We set $\lambda=0.05$ to weight the loss from the foreground and the background.
The overall loss is
\begin{equation}
Loss = \sum_t{\left(Loss^t_{\mathbf{S}} + Loss^t_{\mathbf{L}}\right)} .
\end{equation} 

\subsection{Pose Parsing}
Pose parsing aims to connect the candidate joints and form the full pose. 
Assume that we have an approximate bounding box for each person. 
We perform pose parsing according to Algorithm~\ref{pose_parsing} to get an initial pose for each bounding box.

Note that there possibly exists redundant poses, because (i) one person or part of this person may be visible in multiple bounding boxes, (ii) human detectors may generate redundant detections, which in turn lead to redundant poses.
\begin{algorithm}
	\small
	\caption{Human pose parsing algorithm}\label{pose_parsing}
	\begin{algorithmic}[1]
		\footnotesize
		\For{each bounding box $B_i$ \textbf{in} bounding box set}
			\State calculate confidence for all possible connections by Eq.~\ref{equation:score_connection}
			\For{each limb $l_j$ in DFS order (as shown in Fig.\,\ref{fig:skeleton})} 
				\State $L \gets \{\}$ ~~~// L is the list of candidate connections 
				\For {each possible connection $C_k$ of limb $l_j$ in decreasing order  \\~~~~~~~~~~~~ of confidence} 
					\If{$C_k$ \textbf{not} share joint with connections in $L$}
						\State add $C_k$ into $L$
					\EndIf
				\EndFor
				
				\For{each candidate connection $C_k$ \textbf{in} $L$}
					\State $J_0 \gets$ the start joint of $C_k$
					\State $J_1 \gets$ the end joint of $C_k$
					\If{$J_0$ \textbf{not in} any person's joint list}
						\State creat a new person's joint list, and put $J_0$ and $J_1$ into it
					\Else 
						\State put $J_1$ into the person's joint list which $J_0$ belongs to
					\EndIf
				\EndFor
			\EndFor
		\EndFor
	\end{algorithmic}	
\end{algorithm}

\subsection{Pose NMS}

\label{(sec:pose-nms)}
We perform pose non-maximum suppression (NMS) to detect and remove the redundant poses. Firstly, select the most confident pose as the reference pose $\mathbf{Y}'$, then the other poses close to $\mathbf{Y}'$ are subject to elimination by applying an elimination criterion. Repeat this process on all of the poses set, until at most one unique pose remains in one bounding box.

{\bfseries Pose confidence.} We define the pose confidence by taking into account the covering area of the pose, the confidence of the joints, and the confidence of the connections. Consider a pose $\mathbf{Y}$ with $J$ joints: $\mathbf{Y} = (\mathbf{Y}_{1},\ldots,\mathbf{Y}_{J})$, where $\mathbf{Y}_j$ is the location of joint $j$. Let $s_1(\mathbf{Y})$ and $s_2(\mathbf{Y})$ are the arithmetical average confidence score of the joints and the connections for pose $\mathbf{Y}$. Then we define the confidence of pose $\mathbf{Y}$  as
\begin{equation}
Conf(\mathbf{Y}) = \alpha s_1(\mathbf{Y}) + \beta s_2(\mathbf{Y}) + \gamma \frac{S(B'(\mathbf{Y}))}{S(B(\mathbf{Y}))} ,
\end{equation}
where $B'(\mathbf{Y})$ is the minimum bounding box of  $\mathbf{Y}$, $B(\mathbf{Y})$ is the bounding box of $\mathbf{Y}$, and $S(\cdot)$ denotes the area of the bounding box. We set $\alpha=0.2,\beta=0.2,\gamma=0.6$ as the weight.

{\bfseries Elimination criterion.} In order to eliminate the poses close to the reference pose $\mathbf{Y}'$, we define a distance metric $d(\mathbf{Y},\mathbf{Y}')$ to measure the pose similarity,
\begin{equation}
d(\mathbf{Y},\mathbf{Y}') = \frac{\sum_{j=1}^J(\mathbf{Y}_{j} \neq \mathbf{Y}'_{j})}{max(n_\mathbf{Y}, n_{\mathbf{Y}'})}
\end{equation}
where $n_\mathbf{Y}$ and $n_{\mathbf{Y}'}$ count the number of visible joints in $\mathbf{Y}$ and $\mathbf{Y}'$, $d(\mathbf{Y},\mathbf{Y}')$ computes the percentage of unmatched joints between $\mathbf{Y}$ and $\mathbf{Y}'$.
We specify a threshold $\eta$ as the elimination criterion,
\begin{equation}
f(\mathbf{Y},\mathbf{Y}'|\eta) = \mathbf{1}(d(\mathbf{Y},\mathbf{Y}') \le \eta)
\end{equation}
if $d(\cdot)$ is smaller than $\eta$, the output of $f(\cdot)$ is 1, which indicates that pose $\mathbf{Y}$ should be eliminated due to redundancy with respect to the reference pose $\mathbf{Y}'$.

\subsection{Pose Completion}
Pose completion aims to associate disconnected joints caused by truncation or heavy occlusion to the corresponding pose. Motivated by single-person pose estimation methods, we take a very simple rule: for each missing joint in pose $\mathbf{Y}$, we find a point with the highest confidence in the corresponding cropped confidence map. 
If this point has already been associated to another pose, we find the next highest point constantly until it has not been associated to any other poses. Then we add this point to pose $\mathbf{Y}$. Results show this simple strategy handles disconnected points effectively (Fig. \ref{fig:comparision}, Table \ref{table:dilation_components_AIC_val}).

\section{Experiment}
\label{(sec:experiment)}
\begin{figure}
	\centering
	\includegraphics[width=0.495\textwidth]{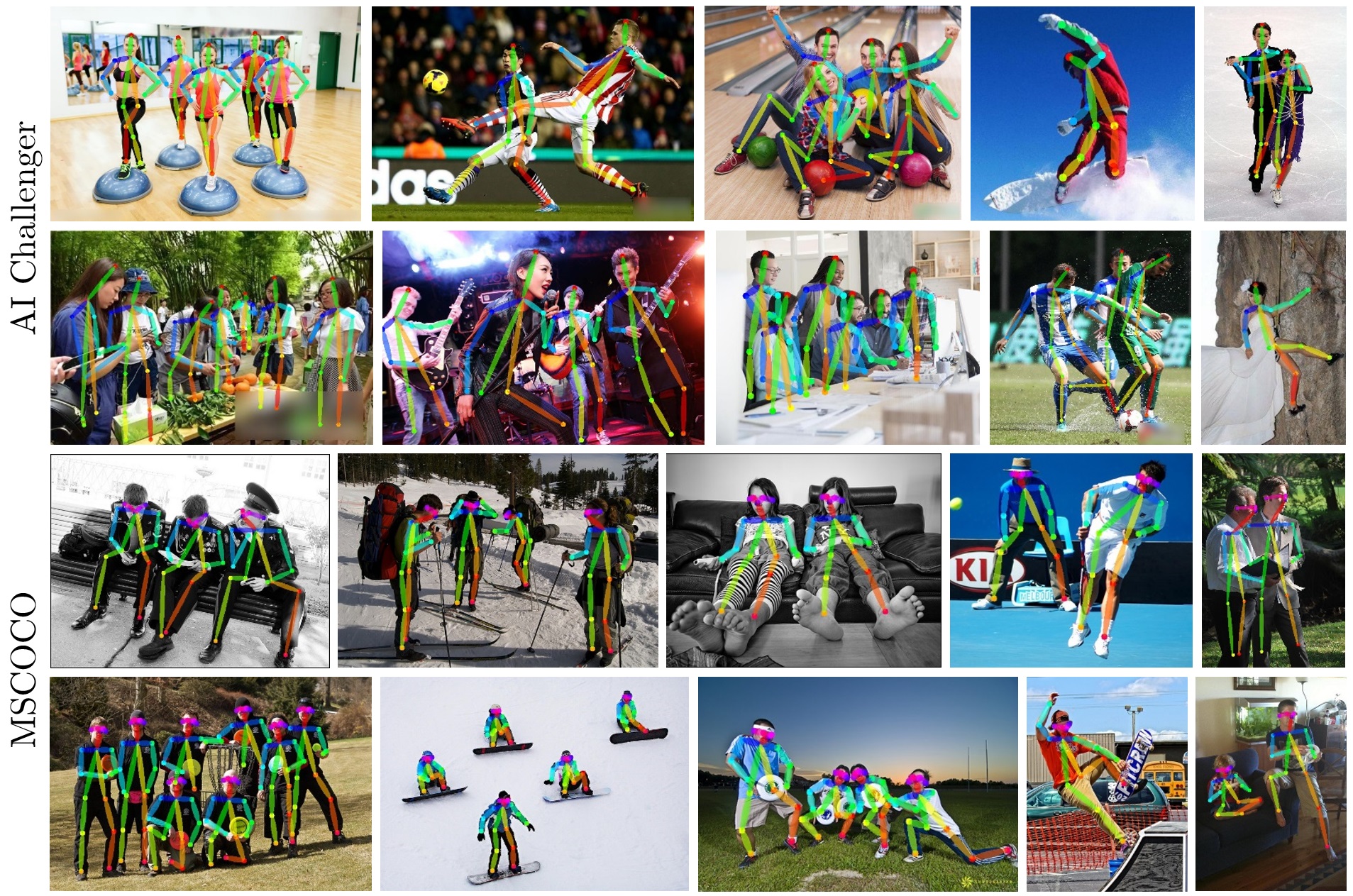}
	\caption{Illustrative results from our method on AI Challenger and MSCOCO datasets. Our method can handle images containing viewpoint and appearance variation, occlusion, crowding, and non-standard pose.}
	\label{fig:result_test}
\end{figure}

We evaluate our method on two standard multi-person pose estimation benchmark dataset qualitatively and quantitatively: (i) AI Challenger keypoints dataset \cite{wu2017ai}, (ii) MSCOCO keypoints dataset \cite{lin2014microsoft}.
Fig.\,\ref{fig:result_test} shows some qualitative results from our method. Both two benchmarks define the Object Keypoint Similarity (OKS) and use the Average Precision (AP) over 10 OKS thresholds as the main competition metric. We refer readers to their website for details\footnote{AI Challenger: \url{https://challenger.ai/competition/keypoint/subject}}$^,$\footnote{MSCOCO: \url{http://cocodataset.org/\#keypoints-eval}}.

\subsection{Implementation Details}
During training, we augment the training data by randomly rotating the input image with a degree in $[-30^{\circ}, 30^{\circ}]$ and flipping it horizontally. We implement our method in the Caffe framework \cite{jia2014caffe}. The model is initialized by the pretrained ResNet50\cite{he2016deep} 
and trained by the Adam solver\cite{kingma2014adam} with the initial learning rate of $8\times10^{-5}$ and the batch size of 16. 

During testing, we use YOLOv2 \cite{redmon2016yolo9000} as the human detector.
In order to guarantee the entire person region can be extracted, detected human proposals are extended by $10\%$ along both the vertical and the horizontal directions. One input image is run at multiple scales ($0.7, 1.0, 1.3$), then the output confidence maps and direction fields are averaged across the scales.
\subsection{Main Results}
\subsubsection{Results on AI Challenger Keypoints Dataset} 
\begin{table}
	\begin{center}
		\caption{Comparison of AP (in \%) on AI Challenger validation set}
		\label{table:result_AIC_val}
		\begin{tabular}{l|c|c|c|c|c}
			\hline 
			& $\mathbf{AP}$ & AP$^{50}$ & AP$^{75}$ & AP$^{M}$ & AP$^{L}$\\
			\hline
			CPM ~\cite{wei2016convolutional} & 62.25 & 93.43 & 68.79 & 48.50 & 64.09\\
			CMU-Posing ~\cite{cao2016realtime} & 59.14 & 88.08 & 64.17 & 44.41 & 61.15\\
			Ours & $\mathbf{66.03}$ & $\mathbf{94.84}$ & $\mathbf{72.93}$ & $\mathbf{52.88}$ & $\mathbf{67.80}$\\
			\hline
		\end{tabular}
	\end{center}
\end{table}

This is the largest dataset for 2D human pose estimation by far, including 300K images and over 700K person instances with skeletal keypoints annotations.

To demonstrate the effectiveness of our method, we compare our method with the typical top-down method taking CPM \cite{wei2016convolutional} as the single-person pose estimator and the bottom-up method proposed by Cao et al. \cite{cao2016realtime} in Table~\ref{table:result_AIC_val}. 
Results show that our method outperforms both the top-down and the bottom-up methods by a large margin. 
For AP$^{50}$ (OKS = 0.5), our method is slightly better than the top-down method. But for AP$^{75}$ (OKS = 0.75), our method gains a far better performance, which illustrates that the pose predicted by our method is more precise. 
It mainly benefits from that the connection relationships between joints are considered. For all three methods, the accuracy for large scale people (AP$^{L}$) is greatly higher than medium scale people (AP$^{M}$), so how to handle medium even small scale people well is still a challenge.

To explain how our method outperforms the top-down and the bottom-up methods, we visualize some predicted poses in Fig.\,\ref{fig:comparision}. Compared with the top-down method, firstly, as shown in  Fig.\,\ref{fig:comparision}\,(a-b), our method is more robust to the shift and the tightness of the body bounding boxes. Secondly, when two people are very close, as shown in Fig.\,\ref{fig:comparision}\,(c-d), the single-person pose estimator fails to determine which person should be annotated, but our method works well. Thirdly, our method performs network feed-forwarding only once, yielding less inference time. Compared with the bottom-up method, our method handles disconnected joints well, and is more robust to truncated or heavily occluded joints (Fig.\,\ref{fig:comparision}\,(a)\,(c-d)). Besides, our method effectively avoids mistake propagation across different poses (Fig.\,\ref{fig:comparision}\,(b-c)\,(e)).

\subsubsection{Results on MSCOCO Keypoints Dataset}
\begin{figure}[tbp]
	\centering
	\subfloat[]{\includegraphics[width=0.15\textwidth]{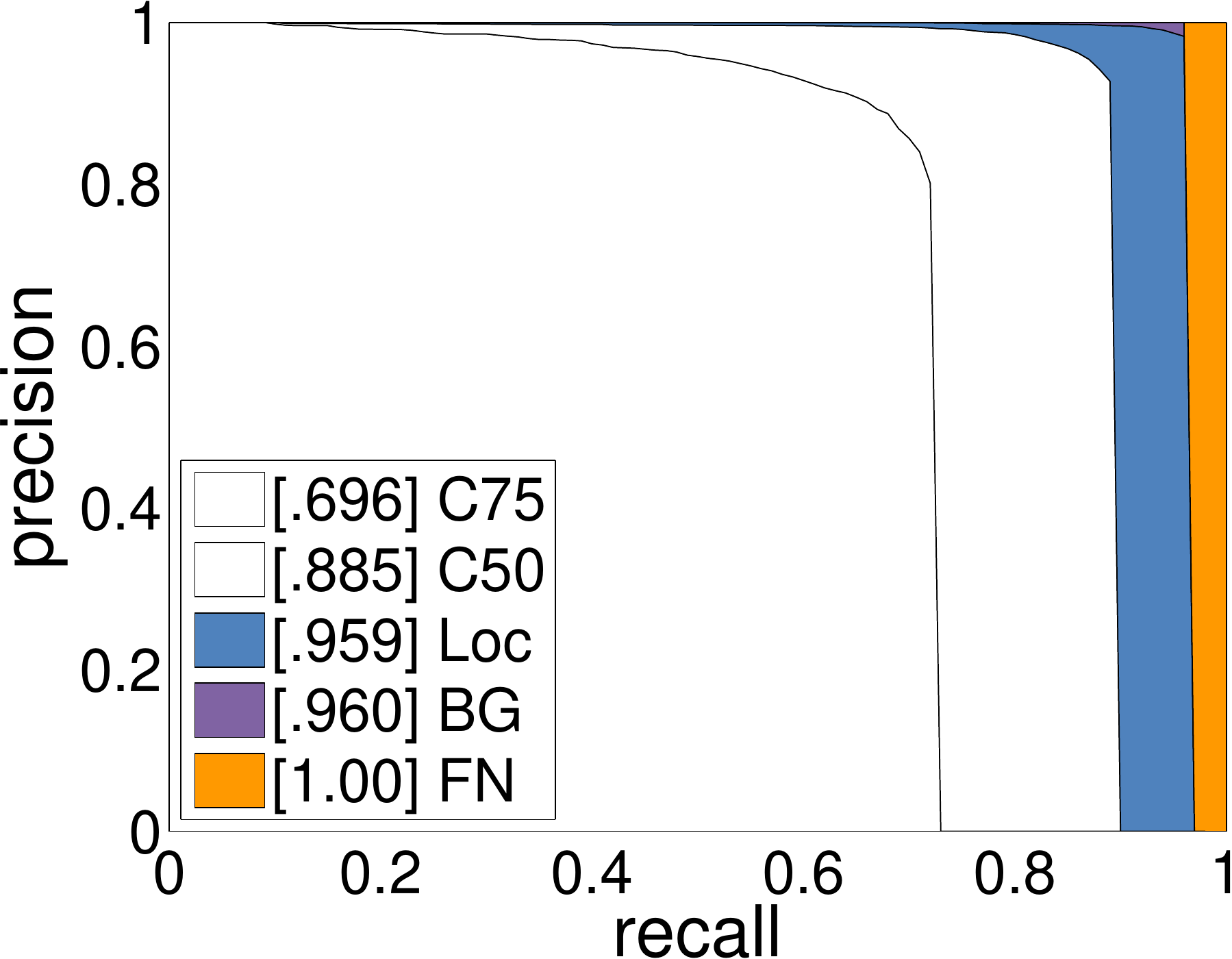}}\
	\subfloat[]{\includegraphics[width=0.15\textwidth]{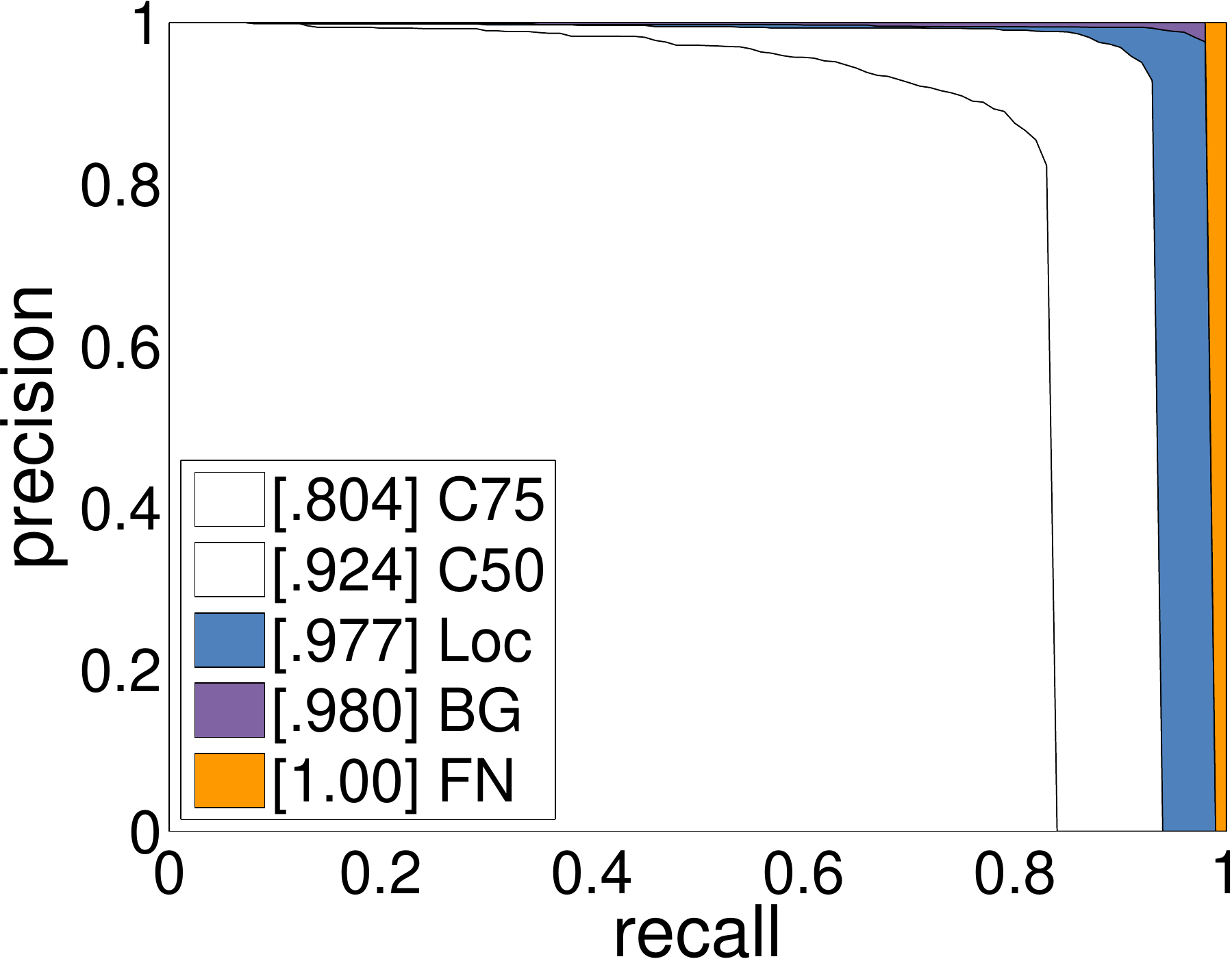}}\
	\subfloat[]{\includegraphics[width=0.15\textwidth]{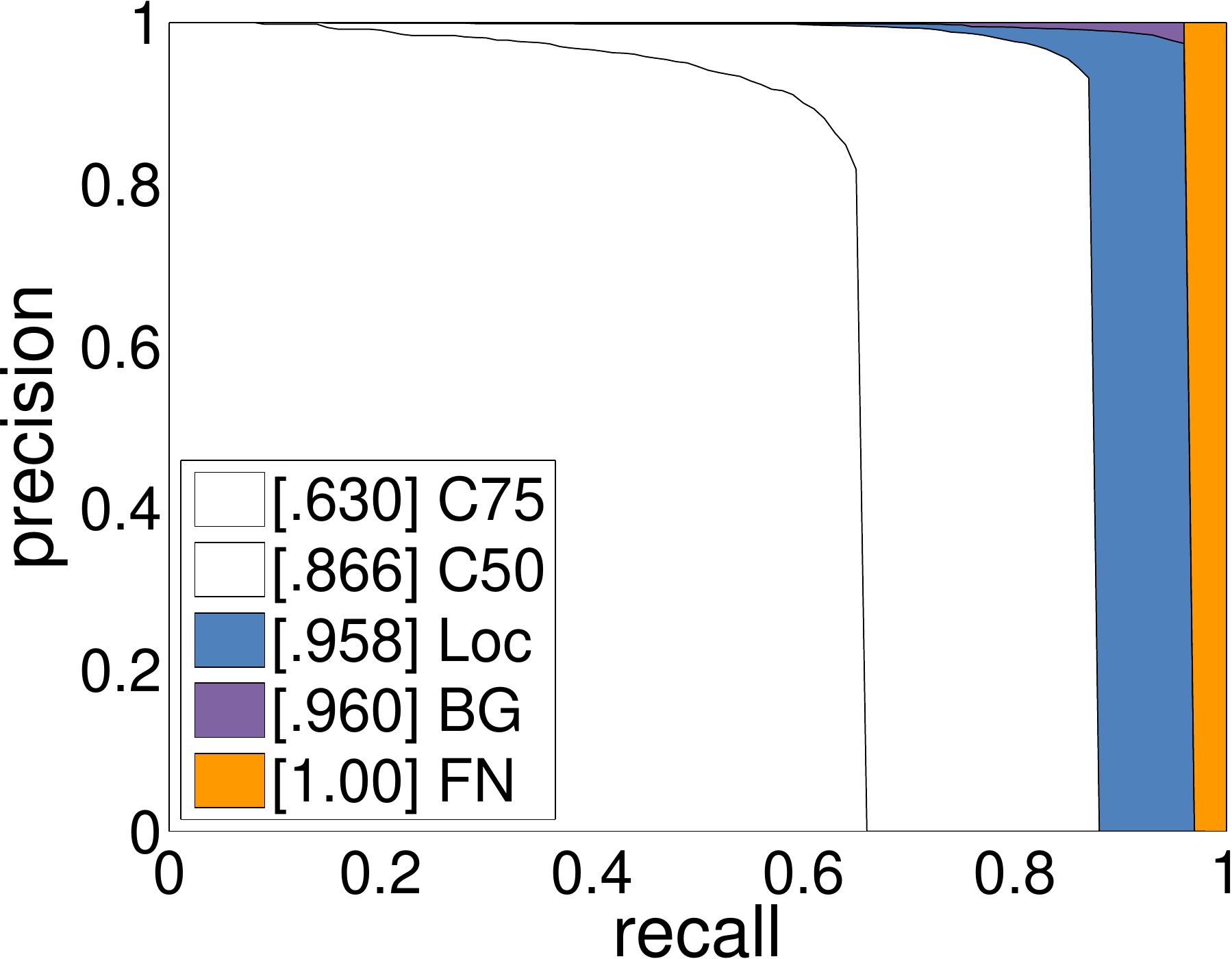}}\\
	\caption{AP performance on MSCOCO validation set. (a), (b) and (c) show the Precise Recall (PR) curve for all people, large scale people and medium scale people respectively. The area under each curve (AUC) is shown in brackets in the legend.}
	\label{fig:PR_curve}
\end{figure}

Table~\ref{table:result_COCO_val} reports the quantitative results compared with the top-down method and the bottom-up method. Results show that our method achieves the best performance.
Fig.\,\ref{fig:PR_curve} shows a breakdown of the errors in our method on the validation set. The plot summarizes the impact of different errors on the performance. Taking the PR curve for all people as an example (Fig.\,\ref{fig:PR_curve} (a)), the overall AP at OKS = 0.75 is 0.696, and the perfect localization would increase AP to 0.959 greatly. Removing all of the background false positive only results in an improvement of about $0.1\% $ (0.960). This indicates that most of the error comes from the imprecise localization, rather than the background confusion.

\begin{table}
	\begin{center}
		\caption{Comparison of AP (in \%) on MSCOCO validation set}
		\label{table:result_COCO_val}
		\begin{tabular}{l|c|c|c|c|c}
			\hline 
			Team & $\mathbf{AP}$ & AP$^{50}$ & AP$^{75}$ & AP$^{M}$ & AP$^{L}$\\
			\hline
			CPM ~\cite{wei2016convolutional} & 62.7 & 86.0 & 69.3 & 58.5 & 70.6\\
			CMU-Posing ~\cite{cao2016realtime} & 58.4 & 81.5 & 62.6 & 54.4 & 65.1\\
			Ours & $\mathbf{64.7}$ & $\mathbf{88.5}$ & $\mathbf{69.6}$ & $\mathbf{59.4}$ & $\mathbf{73.7}$\\
			\hline
		\end{tabular}
	\end{center}
	$*$ Except for ours, all other results are obtained from the original paper \cite{cao2016realtime}.
\end{table}
\subsubsection{Runtime Analysis}
We test the runtime of our method on a PC equipped  with an Intel Core i7-7700k CPU, and a NVIDIA GeForce GTX-1080Ti GPU. We test our method using videos with a varying number of people. All videos are resized to $848 \times 480$. The average network feed-forwarding time per frame is 63 ms, and the average pose parsing time per frame is 24 ms. Our method has achieved 11.43 fps for a video. Overall, our approach keeps the advantage of the bottom-up approaches which perform network feed-forwarding only once per image, so the runtime increases relatively slowly with the increasing number of people. 
\begin{table}
\begin{minipage}{0.5\textwidth}  
	\begin{center}
		\caption{Analysis on different network structures}
		\label{table:dilation_models_AIC_val}
		\begin{tabular}{l|c|c|c|c}
			\hline 
			Network Structures& AP& $\Delta$ & Forward& Param\\
			 &(\%) &(\%) & Time (ms) & Size \\
			\hline
			6-stage + VGG & 62.25 & - & 123.54 & 51.29 M\\
			6-stage + ResNet & 64.81& $+2.56$ & ~80.77 & 51.97 M\\
			3-stage + ResNet & 63.32 & $+1.07$ & ~43.67 & 26.00 M\\
			3-stage + ResNet + 2$\times$ 
			& 66.03 & $+3.78$ & ~63.19 & 27.42 M\\
			6-stage + ResNet + 2$\times$ & 66.12 & $+3.87$ & 126.88 & 54.73 M\\
			\hline
		\end{tabular}
	\end{center}
	$*$ 2$\times$: upsample the outputs by 2 times using deconvolution layers.
\end{minipage}
\end{table}

\subsection{Ablation Study}
In this section, we analyze the functions of the four components in our approach. We conduct the experiments on the validation set of AI Challenger.

\subsubsection{Network Structures}
We experiment on five different network structures considering three factors: (i) sub-network for feature extraction; (ii) upsample the network outputs by the deconvolution layers; (iii) the number of cascaded stages. Our baseline is the 6-stage network with the VGG19 feature extraction module and without deconvolution layers. As shown in Table~\ref{table:dilation_models_AIC_val}, after replacing the VGG19 with the ResNet50, the AP increases 2.56$\%$. If fewer stages (3-stage) are cascaded, with fewer parameters and less network feed-forwarding time, the AP drops to 63.32$\%$. Generally speaking, there is a trade-off between accuracy, speed as well as parameters size. After introducing deconvolution layers to upsample the outputs, without incurring significant extra computation cost, the AP increases 2.71$\%$ with a 3-stage network. However, we find the improvement is not significant after increasing the stage number to 6 or more. It demonstrates that the 3-stage network, with residual feature extraction module and deconvolution layers, achieves much better trade off between accuracy, speed and parameters size.

\subsubsection{Pose Parsing Methods}
Compared with the pose parsing method of \cite{cao2016realtime}, Table~\ref{table:dilation_components_AIC_val} shows that our method outperforms by a large margin ($+6.89\%$). It mainly benefits from that our method avoids the disconnected parts of poses and the mistakes propagation across different poses effectively after introducing the bounding box constraint.

\subsubsection{Pose NMS}
Table~\ref{table:dilation_components_AIC_val} shows that the AP drops significantly ($-2.11\%$) if pose NMS is removed. The reason is that the number of redundant poses will ultimately decrease the final average precision.

\subsubsection{Pose Completion}
As shown in Table~\ref{table:dilation_components_AIC_val}, after removing pose completion, the AP drops to $64.34\%$. It illustrates that our method handles considerable disconnected joints effectively, which are abandoned in the bottom-up method\cite{cao2016realtime}. 

\begin{table}  
	\begin{center}
		\caption{Analysis on the function of each component in our approach}
		\label{table:dilation_components_AIC_val}
		\begin{tabular}{l|c|c}
			\hline 
			Methods & AP(\%) & $\Delta$ (\%) \\
			\hline
			pose parsing by \cite{cao2016realtime} & 59.14 & $-6.89$ \\
			without pose NMS & 63.92 & $-2.11$\\
			without pose completion & 64.34 & $-1.69$ \\
			our baseline & $\mathbf{66.03}$ & -\\
			\hline
		\end{tabular}
	\end{center}
\end{table}

\section{Conclusion}

In this paper, we present a new bottom-up approach to the multi-person pose estimation problem. The main idea is using a residual network to learn both the confidence maps of joints and the connection relationships between joints, and then take advantage of the body bounding box to parse the pose. In contrast to the top-down methods, our method takes less inference time, since it performs network feed-forwarding only once per image. Secondly, our method is more robust to body bounding box shift and tightness. What's more, in our method, the network does not infer which person should be annotated when multiple persons are visible in one bounding box, so our method handles images with crowding better. In contrast to the bottom-up method\cite{cao2016realtime}, our method avoids mistake propagation across different poses, and addresses disconnected points effectively. 
Results showed that our method achieves a better performance both in the accuracy and the runtime. In our future work, it would be interesting to explore the possibility of training our network with the human detector in an end-to-end manner.

\section*{Acknowledgment}
The authors would like to thank the anonymous reviewers for their valuable comments. This work was partially supported by National Key Research and Development Program of China (No. 2016YFB1001501) and NSFC (No.61379068).

\bibliographystyle{IEEEtran}
\bibliography{IEEEabrv,reference}
%



\end{document}